\documentclass[sigconf]{acmart}
\makeatletter
\def\@ACM@checkaffil{
    \if@ACM@instpresent\else
    \ClassWarningNoLine{\@classname}{No institution present for an affiliation}%
    \fi
    \if@ACM@citypresent\else
    \ClassWarningNoLine{\@classname}{No city present for an affiliation}%
    \fi
    \if@ACM@countrypresent\else
        \ClassWarningNoLine{\@classname}{No country present for an affiliation}%
    \fi
}
\makeatother

\usepackage{ulem}
\usepackage{placeins}
\usepackage[utf8]{inputenc} 
\usepackage[T1]{fontenc}    
\usepackage{hyperref}       
\usepackage{url}            
\usepackage{booktabs}       
\usepackage{amsfonts}       
\usepackage{nicefrac}       
\usepackage{microtype}      
\usepackage{xcolor}         
\usepackage{soul}
\usepackage{cleveref}
\usepackage{graphicx}
\usepackage{makecell}
\usepackage{soul}
\usepackage{multirow}
\usepackage[title]{appendix}
\usepackage{tablefootnote}

\newcommand{\sysname}{$\mathsf{Fieldy}$} 

\title[Fine-grained Attention in Hierarchical Transformers for Tabular Time-series]{
Fine-grained Attention in Hierarchical Transformers\\for Tabular Time-series} 
\renewcommand\footnotetextcopyrightpermission[1]{} 

\copyrightyear{} 
\acmYear{2024} 
\setcopyright{none} 
\acmConference[KDD'24 -- 10\textsuperscript{th} MiLeTS Workshop]{10\textsuperscript{th} Mining and Learning from Time Series Workshop (MiLeTS)}{August 26, 2024}{Barcelona, Spain}
\acmBooktitle{ACM SIGKDD'24 -- 10\textsuperscript{th} Mining and Learning from Time Series Workshop (MiLeTS), August 26, 2024, Barcelona, Spain}
\acmDOI{}
\acmISBN{}

\author{Raphael Azorin}
\affiliation{%
  \institution{EURECOM, Huawei Technologies}
  }
\country{}
\email{last@eurecom.fr}  

\author{Zied Ben Houidi}
\affiliation{%
  \institution{Huawei Technologies}
  }
\country{}
\email{first.last@huawei.com}  

\author{Massimo Gallo}
\affiliation{%
  \institution{Huawei Technologies}
  }
\country{}
\email{first.last@huawei.com}  
  
\author{Alessandro Finamore}
\affiliation{%
  \institution{Huawei Technologies}
  }
\country{}
\email{first.last@huawei.com}  

\author{Pietro Michiardi}
\affiliation{%
  \institution{EURECOM}
  }
\country{}
\email{last@eurecom.fr}

\newcommand{\replaceok}[2]{{#2}}

\begin{document}

\begin{abstract}
  Tabular data is ubiquitous in many real-life systems. In particular, time-dependent tabular data, where rows are chronologically related, is typically used for recording historical events, e.g., financial transactions, healthcare records, or stock history. 
  Recently, hierarchical variants of the attention mechanism of transformer architectures have been used to model tabular time-series data.
  At first, rows (or columns) are encoded separately by computing attention between their fields. Subsequently, encoded rows (or columns) are attended to one another to model the entire tabular time-series. While efficient, this approach  constrains the attention granularity and limits its ability to learn patterns at the field-level across separate rows, or columns.
  We take a first step to address this gap by proposing \sysname{}, a fine-grained hierarchical model that contextualizes fields at both the row and column levels.
  We compare our proposal against state of the art models on regression and classification tasks using public tabular time-series datasets. Our results show that combining row-wise and column-wise attention improves performance without increasing model size. 
  Code and data are available at \url{https://github.com/raphaaal/fieldy}.
\end{abstract}

\maketitle

\vspace{-1mm}
\section{Introduction}

Sequential tabular data is widely used in the industry to represent financial transactions recorded in a bank database \cite{luetto2023one}, medical records stored by a hospital \cite{zhang2020time} or customers purchase history maintained in a CRM system \cite{zhang2019deep}, to name a few examples.
Such tabular data are composed of rows and columns, each row corresponding to a record which collects values for each column. Different from traditional multivariate time-series, tabular time-series often present categorical variables. 
Unlike classic tabular data which considers separate rows as distinct input samples 
for a given downstream task,
records in sequential tabular data span 
multiple rows, a property that can be 
exploited
when time-related fields are present (see \Cref{tabular_timeseries_data}). 
Common examples of tabular time-series tasks take multiple rows in input and provide some prediction, 
e.g.,
detecting fraud from sequences of financial transactions \cite{padhi2021tabular}, predicting click-through rate from past online behavior \cite{muhamed2021ctr} or forecasting pollution from historical data \cite{padhi2021tabular}.

As tabular time-series tasks are intrinsically sequential, the research community started to investigate how to leverage the success of 
transformer architectures {\cite{vaswani2017attention}} from Natural Language Processing (NLP) within the tabular domain {\cite{padhi2021tabular, luetto2023one, fata}}. 
In a nutshell, by exploiting the attention mechanism, a transformer can relate the tokens that compose a sequence to one another, hence learning relationship patterns between time-steps. 
This mechanism led to significant improvements in NLP tasks, such as sequence classification (e.g., sentiment analysis), token classification (e.g., entity recognition) or sequence regression (e.g., emotion level prediction). 
Just as transformers extract rich features from sequences of words, the sequence of records in a table is crucial for extracting meaningful patterns in tabular data modeling.

\begin{figure}[t!]
\centering
\includegraphics[width=0.90\linewidth]{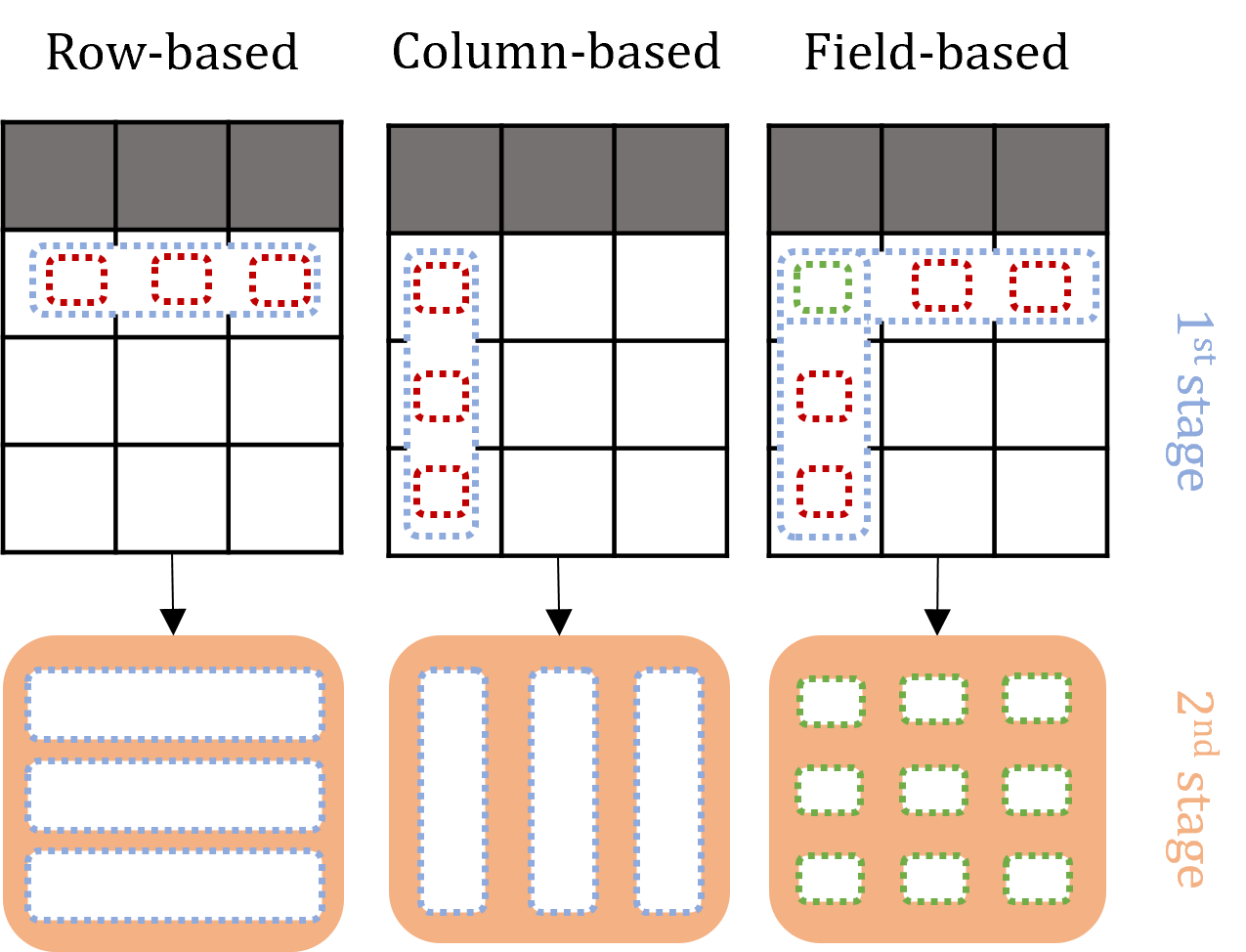}
\caption{Hierarchical transformers schematic view.
}
\label{archs_intro}
\vspace{-4mm}
\end{figure}

As tabular data is bi-dimensional, and both rows and columns carry semantics, 
the transformer architectures available in literature for tabular time-series often present a hierarchical design. In a first stage, each row (or column) is encoded separately by aggregating the outputs of a transformer computing attention across its fields, as shown in \Cref{archs_intro} -- left and center. 
In the second stage, these encoded rows (or columns) are then passed to another transformer. 
The result of the second stage is the final encoding of the entire tabular time-series, 
which is typically processed by 
additional fully connected layers that specialize the model
for solving a given downstream task, e.g., sequence classification.
From a tabular perspective, this two-stage process first captures interactions between fields within \replaceok{each row (resp. column), and then interactions between rows (resp. columns) within each time-series.}{a given dimension (row or column), and then interactions among those representations.}

While hierarchical architectures capture all table dimensions, 
they don't do that \textit{simultaneously}, hence
limiting visibility on more subtle cross-field relationships important for the downstream task.
\textcolor{black}{In Appendix \ref{app:field_attention}, we empirically demonstrate that this shortcoming hinders learning interactions between fields across separate rows, due to the coarse-grained aggregation performed in the second stage.}
This suggests that a \textit{field-wise attention} mechanism can be an appealing alternative to capture the intricate relationships \textit{between all the fields across the full tabular time-series}, as illustrated in {\Cref{archs_intro} -- right.}
In this paper, we implement this mechanism, introducing \sysname{}, a novel architecture that combines row-wise and column-wise transformers in the first stage to learn field representations. 
These contextualized field representations are then merged, reshaped and passed to the second-stage transformer to produce the final encoding of the entire tabular time-series. 
Consequently, \sysname{} enables fine-grained attention across \textit{all} the fields composing a tabular time-series, while incorporating row-level and column-level information.
We compare our solution against both state of the art transformer architectures and Machine Learning (ML) tree-based ensemble algorithms
using two popular tabular datasets and show that \sysname{} outperforms alternative methods. 

The remainder of this paper is structured as follows. First, we review prior work on transformers for standard tabular data and tabular time-series (\Cref{related_work}). Then, we present \sysname{} and highlight its differences with row-based and column-based hierarchical transformers (\Cref{methodology}) before presenting our evaluation protocol and results (\Cref{experiments}). Last, we discuss our findings and identify areas for future work (\Cref{discussion}). 

\begin{table}
\small
    \centering
    \caption{A tabular dataset. 
    Records may be grouped by \texttt{Patient} to produce tabular time-series.
    } 
    \label{tabular_timeseries_data}
    \begin{tabular}{ |c| c| c| c| c| c| }
     \hline
     \textbf{Timestamp} & \textbf{Patient} & \textbf{Disease} & \textbf{Therapy}
     & \textbf{Temperature} 
     \\
     \hline
     2024-06-01 & 012 & Tuberculosis & A 
     & 38.2 \\ 
     2024-01-15 & 012 & Flu & B 
     & 38.3 \\  
     2023-12-28 & 456 & Hemophilia & C 
     & 37.5 \\  
     2023-07-26 & 012 & Angina & B 
     & 37.9\\
     2023-01-28 & 456 & Sinusite & B 
     & 37.3 \\    
     2022-02-11 & 789 & Flu & D 
     & 38.1 \\ 
     \hline
    \end{tabular} 
\end{table}

\section{Related work} \label{related_work}

In this section, we first review prior work on Deep Learning (DL) for standard tabular data. We then focus on the specific case of tabular time-series, using {\Cref{tabular_timeseries_data}} as toy example. Finally, we discuss how our contribution fits within the existing literature.

\paragraph{Transformers for tabular data}
In traditional tabular data modeling, each row is treated as a distinct input sample for which a prediction or a classification needs to be made. A variety of proposals address this type of data using both DL and ML methods.
For instance, a recent study~{\cite{grinsztajn2022tree}} explores DL approaches, including CNNs and transformers, as well as ML tree-based models like XGBoost and Random Forests, applied to tabular data. 
Although the study shows that gradient-boosted trees outperform deep learning algorithms on most datasets, FT-Transformer \cite{gorishniy2021revisiting} emerges as a promising architecture. In particular, FT-Transformer encodes each row by computing attention between its fields and then processes it, with a final fully connected layer. 
FT-Transformer relies on feature tokenizers to embed both categorical and numerical features that may appear in tabular data, which is an uncommon and challenging scenario for ML approaches.
Note that this method is row-based but not hierarchical as each record is encoded separately. 
Alternatively, Tabbie \cite{iida2021tabbie} proposes to encode each table field by averaging representations of its row and its column. To do so, two transformers are tasked to encode each row and each column separately, in order to form ``contextualized'' fields representations by averaging their intersections. Once the full table has been encoded, each row is processed by a final fully connected layer.
Note that Tabbie is not a hierarchical architecture, as it only operates at the field-level granularity, either contextualized by row or by column.
While Tabbie is not adapted to tabular time-series, as it does not consider subsets of time-dependent rows, it is the closest to our proposal in the way it attends to row and column contexts.

\begin{table}[!b]
\small
\centering
\caption{Transformer-based models comparison.}
\label{tab:lit_review}
\begin{tabular}{
@{}l
@{$\,\,\,$}c 
@{$\,\,\,$}c 
@{}
}
\toprule
\textbf{Model} & \textbf{\makecell{Architecture}} & \textbf{Attention axis}  \\ 
\midrule
FT-Transformer \cite{gorishniy2021revisiting} & Single-stage & Horizontal  \\
Tabbie \cite{iida2021tabbie}& Single-stage & Horizontal \& vertical  \\
\midrule
TabBERT \it{(row-based)} \cite{padhi2021tabular}  & Two-stage & Horizontal $\rightarrow$ Vertical  \\
TabBERT \it{(column-based)}  & Two-stage & Vertical $\rightarrow$ Horizontal  \\
\sysname{} \it{(ours)} & Two-stage & Horizontal \& vertical $\rightarrow$ Fields \\
\bottomrule  
\end{tabular}
\end{table}

\paragraph{Transformers for tabular time-series}
In their most prevalent form, tabular time-series are 
a specific family of tabular data where records are interdependent and ordered or timestamped (e.g., with an explicit {{\tt Timestamp}} field as in {\Cref{tabular_timeseries_data}}), often presenting an ``entity'' identifier 
used for grouping records into samples of interest. 
For example, {\Cref{tabular_timeseries_data}} yields three distinct tabular time-series corresponding to three patients' history, each of which could be the input to a machine learning model. 
In this case, a classic ML algorithm would simply take as input the concatenation of multiple records, hence returning to the traditional tabular data scenario. 
Conversely, more recent sequence models open the way for using hierarchical approaches to tabular time-series modeling. 
In this context, TabBERT \cite{padhi2021tabular} is composed of a first-stage transformer that encodes each row in a tabular time-series, and a second-stage transformer that processes the encoded rows to generate a full sequence representation. This final representation is then passed to a fully connected layer, e.g., a classification head, to perform the downstream task.
Given its design, we refer to this architecture as ``row-based''.
Alternatively, instead of encoding rows in the first stage, a variation of TabBERT may encode each column separately, and process these encoded columns in a second stage to generate the full tabular time-series representation. We refer to this inversion of TabBERT as ``column-based''.
Several works have extended this hierarchical approach to model tabular time-series, such as \cite{luetto2023one} that extends it to heterogeneous tabular time-series, i.e., with different number of columns, and \cite{fata} that exploits time-deltas to better model time differences in the context of tabular time-series.
Our work builds on hierarchical tabular time-series modeling, extending it with the ability to integrate field interactions across both rows \textit{and} columns \textcolor{black}{simultaneously}. 

\begin{figure*}[!ht]
\centering
\includegraphics[width=0.95\textwidth]{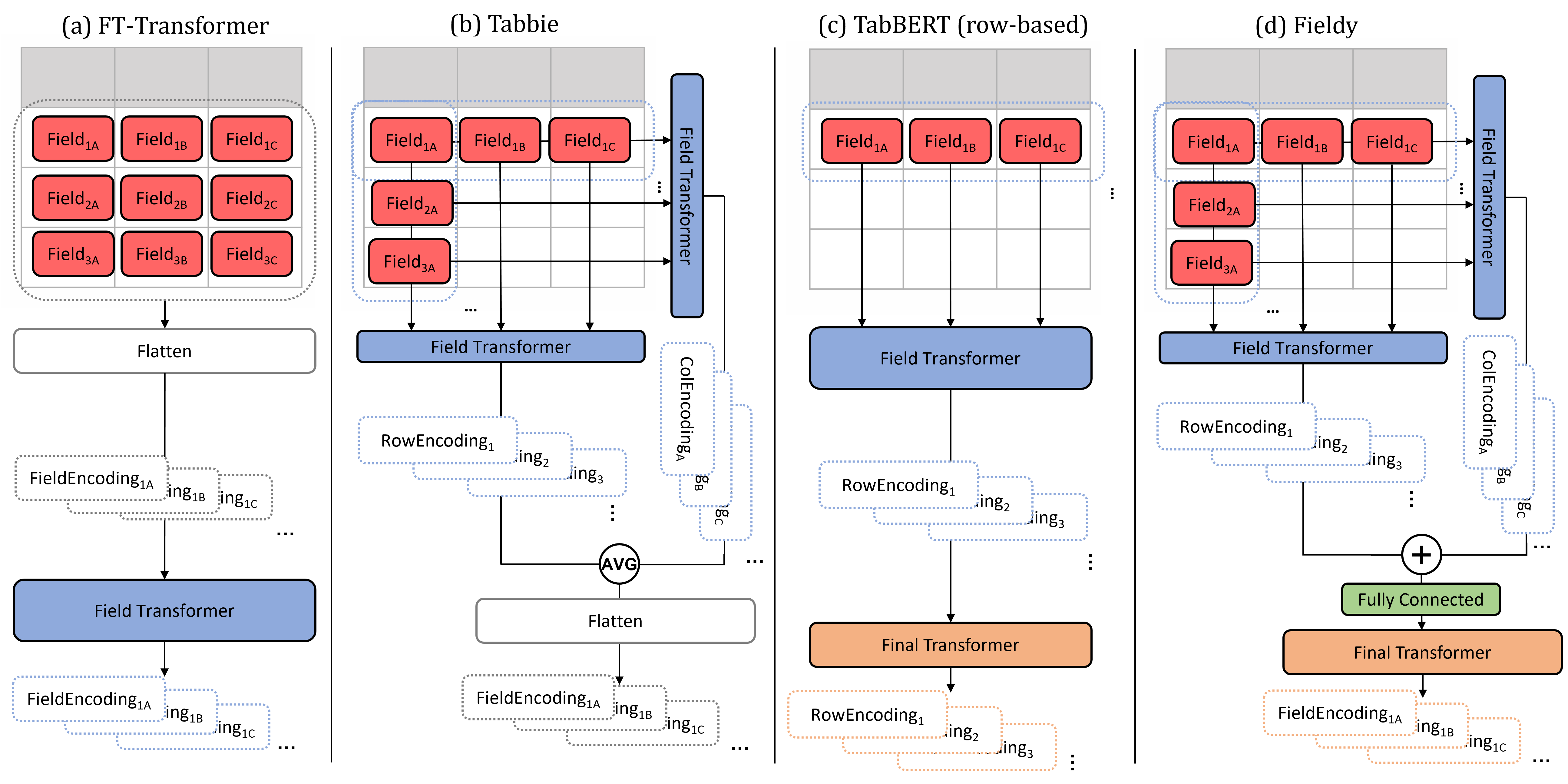}
\caption{Detailed view of transformer architectures for tabular time-series. The \textcircled{+} operator denotes concatenation.}
\label{detail_hierarch_comparison}
\end{figure*}

\paragraph{Our contribution}
In this paper we set out to evaluate whether a \textcolor{black}{hierarchical} model with a finer cross-field attention
provides more appropriate representations of tabular time-series, as opposed to a coarse aggregation of row or column embeddings. 
\textcolor{black}{In a nutshell, we propose an architecture that simultaneously captures row-wise and column-wise interactions in a first stage, to learn contextualized field representations that are related to one another in a second stage.}
To properly isolate this effect and focus the comparison on this particular design choice, we compare this field-based attention approach to its row-based and column-based counterparts in similar conditions. In other terms, in this paper, we do not aim to introduce novel tokenizers or novel pre-training tasks which are orthogonal directions of research. 
Additionally, since our approach relates any field to any other during the second stage, unlike row-based and column-based, we compare it also to a single-stage baseline where all the rows of interest are flattened and fed to \textcolor{black}{a unique} transformer. \textcolor{black}{This approach can be seen as an adaptation of FT-Transformer for tabular time-series}, effectively resulting in a comparison point that links any field to any other. 
Finally, to complete the design space, another interesting comparison point we consider consists in constructing contextualized field embeddings that incorporate both rows and columns and then feeding the flattened sequence to a linear layer. This approach could be seen as a straightforward adaptation of Tabbie to the context of tabular time series. 

\section{Methodology}
\label{methodology}

In this section, we first dive into the details of current approaches and their limitations. We then present our \textcolor{black}{field-based hierarchical approach}
that integrates both row-wise and column-wise interactions. We finally introduce the positional embeddings schemes used to encode table structure.

\paragraph{State of the art limitations}
Tabular time-series modeling approaches considered in this paper 
are summarized in \Cref{tab:lit_review}.
Non-hierarchical baselines such as FT-Transformer and Tabbie are single-stage architectures designed for traditional (i.e., non sequential) tabular data. 
When adapting FT-Transformer to tabular time-series, the input sequence is flattened to form a single long row of repeated features at various time-steps, as depicted in \Cref{detail_hierarch_comparison} (a). As this model computes attention across all the fields composing a single row, it captures relationships between all the fields across all rows and columns as a result of the input flattening. However, the downside of this flattening is that this adaptation of the FT-Transformer model is oblivious to the table structure. We consider this as a baseline transformer attending to all fields.
Alternatively, when adapting Tabbie to ingest tabular time-series, we limit its row-wise contextualization to a subset of the table: the rows composing the input sequence. As it relates each field to other fields present in the same row and column, Tabbie is able to capture relationships along both table axes. However, as depicted in \Cref{detail_hierarch_comparison} (b), this approach is not equipped with a second-stage to relate all the fields outside of their original row and column.

Regarding hierarchical models, recall that two-stage architectures from the literature are either ``row-based'' or ``column-based''. Thus, they condition the attention mechanism between distinct fields along a particular table axis in the first stage, as illustrated in \Cref{detail_hierarch_comparison} (c). 
While these approaches enforce the tabular time-series row (or column) structure, they fail to capture relationships between fields across separate rows (or columns). For instance, 
considering \Cref{tabular_timeseries_data}, a row-based architecture cannot explicitly relate, i.e., by means of attention, a {\tt Disease} field to a {\tt Therapy} field if they belong to different rows, nor can it \textcolor{black}{directly} relate two {\tt Disease} fields that belong to two different rows. \textcolor{black}{As shown empirically in Appendix \ref{app:field_attention}, this limits the ability of hierarchical models from the literature to learn fine-grained relationships at the field level that might be relevant for a given downstream task.}

\paragraph{Field-wise attention}
Two-stage approaches rely on the assumption that the Field transformer in the first stage \replaceok{can}{is sufficiently powerful to} extract \replaceok{sufficiently}{}expressive row/column representations for the Final transformer, \replaceok{not to require to learn}{rather than fostering the learning of} fine-grained field relationships in the second stage. To capture relevant fields interations that may be beneficial to tabular time-series tasks, we introduce \sysname{}: a novel hierarchical transformer that combines both in-row and in-column attention, as well as cross-field attention. 
As depicted in {\mbox{\Cref{detail_hierarch_comparison} (d)}}, we propose a two-stage architecture in which the first stage consists of two Field transformers that operate simultaneously: one is responsible to contextualize each field row-wise, and the other is responsible to contextualize each field column-wise. The resulting encoded rows and columns are then concatenated 
to constitute field representations.
These contextualized cell representations are then passed through a fully connected layer to produce rich representations before being processed by a Final transformer, which would attend to all fields. 
Note that our first stage resembles Tabbie; yet, it differs for two key design choices. First, while Tabbie creates a field embedding through a deterministic simple average of the field's row and column embeddings, \sysname{} \textit{learns} how to combine the two embeddings in the first stage.  
Second, unlike Tabbie which passes the embedded fields to the final fully connected layer (e.g., a classifier head), we adopt a second-stage transformer that relates all the field representations to each other.  
At last, the entire encoded tabular time-series coming from the Final transformer is processed by a fully connected layer, fine-tuned on a specific downstream task (e.g., sequence classification).
In order to fairly compare \sysname{} against row-based and column-based architectures, we reduce the size of its Field transformers to reach the same model size. Nonetheless, we note that \sysname{} requires increased computational effort, an aspect which we discuss in \Cref{discussion}. 

\paragraph{Positional encoding to capture table structure}
Compared to row-based or column-based hierarchical approaches that model the table structure by design, \sysname{} requires additional information.
In the row-based architecture, the Field transformer implicitly establishes the table's horizontal structure by computing attention horizontally across all the fields of the same row. 
The Final transformer then enforces the table vertical structure as it ingests a sequence of per-row representations.
Conversely, in the column-based architecture, the Field transformer is provided with the table's vertical structure, as its input is a sequence of fields from distinct rows. Then, the Final transformer enforces the table horizontal structure by attending between per-column representations.
For these two approaches, it is only the \textit{position} of the rows (or columns) that is unknown from the model, but their delimitation is apparent by design.
On the other hand, the design of \sysname{} does not incorporate the table structure by default, as the Final transformer ingests a long sequence composed of all the contextualized fields. Thus, without additional information, the Final transformer can only access a bag of fields, being oblivious to the original rows or columns they come from. 
Therefore, we incorporate row and column positional embeddings~\cite{devlin2018bert}. Before being passed to the Final transformer, each contextualized field is augmented (by means of element-wise addition) with two embeddings: one carrying its original row position and one carrying its original column index. 
To ensure a fair comparison between architectures, we also add these row position and column index embeddings to all the other models. We discuss the effect of these additions in \Cref{experiments}.

\section{Evaluation}
\label{experiments}

In this section, we first detail the datasets and models configurations considered to compare transformers architectures on tabular time-series. Then, we present the results of our evaluation and expand our analysis with an ablation study.

\subsection{Datasets}

\paragraph{Pollution -- Regression}
The UCI Beijing Pollution Dataset \cite{misc_beijing_multi-site_air-quality_data_501} consists in predicting air pollution particles from 12 sites located in Beijing. This dataset has been used to evaluate row-based hierarchical transformers in \cite{padhi2021tabular, luetto2023one}. It is a multi-regression task taking as input 10 features (such as temperature, pressure, etc.) measured on an hourly basis during 10 time-steps. The labels to predict correspond to the PM\textsubscript{2.5} and PM\textsubscript{10} concentrations for each time-step, i.e., 20 labels for each input sequence.
We replicate the pre-processing described in \cite{padhi2021tabular}, i.e., we discretize numerical variables using 50 quantiles and normalize the targets. 
After data exploration, we choose to include 6 additional features not present in the pre-processing from related work. These engineered features correspond to the measurement site name, the hour of the measurement, the day of the month, the weekday, the month, and the year.
Finally, we remove 4\% of outliers when PM\textsubscript{10} > PM\textsubscript{2.5} as mentioned in \cite{wu2018probabilistic}.
As in prior work, the evaluation metric is defined as the RMSE averaged across the concentration targets. In total, this dataset contains 67K tabular time-series. The pre-training dataset consists of the same dataset excluding the labels (more details on data splits in Appendix \ref{app:datasets}).

\paragraph{Loan default -- Classification}
The PKDD'99 Bank transactions dataset \cite{berka1999pkdd} contains real transaction records for 4,500 clients of a Czech bank. It has been used in \cite{luetto2023one} to evaluate hierarchical transformers. The considered task consists in predicting if a client will default its loan based on his prior transactions. Six input features describe each transaction (amount, type, etc.) and the label is binary, i.e., one for each clients' input sequence.
As for the pollution dataset, we pre-process the data similarly to prior work \cite{luetto2023one}, 
i.e., using 50 quantiles to discretize each numerical feature and splitting the timestamp into three fields: day, month and year. 
Note that we include an additional weekday feature that proved meaningful during data exploration. As in \cite{luetto2023one}, we segregate the data into 3,818 clients with unlabeled transactions for pre-training, and 682 clients with labeled transactions for fine-tuning.
In order to increase the dataset size, we consider any sequence of 10 consecutive transactions for each client, while \cite{luetto2023one} considered only his last 150 transactions.
We thus obtain 5K tabular time-series for fine-tuning, instead of only 682.
The evaluation metric is defined as the Average Precision (AP) score to take into account the class imbalance of this dataset (more details on data splits in Appendix \ref{app:datasets}).

\subsection{Models}

\paragraph{Architectures}
For each dataset, we consider three hierarchical transformers: row-based, column-based and our \replaceok{row \& column-based}{field-based} proposal \sysname{}. We use the official code from TabBERT \cite{padhi2021tabular} to implement row-based and column-based baselines. Additionally, we evaluate two single-stage baselines: FT-Transformer and Tabbie, using our own implementation to adapt them to tabular time-series. We size all models to amount to the same total number of parameters. As \sysname{} requires two Field transformers in its first stage, we reduce their number of layers to make the comparison fair. We keep all the other hyper-parameters related to model capacity (hidden dimensions, number of attention heads, etc.) the same across all models, similar to prior work \cite{padhi2021tabular, luetto2023one, fata}. 
Models are pre-trained for 24 epochs (\textit{Pollution}) or 60 epochs (\textit{Loan default}), and fine-tuned for 20 epochs. The best models are selected based on their score on a validation set and evaluated on a held-out test set.
Additionally, we include two non-deep learning baselines: XGBoost \cite{chen2016xgboost} and a linear model (linear or logistic regression). Note that these two baselines use exactly the same pre-processed input features and labels as the hierarchical transformers, which means that all numerical features are quantized. Both of these baselines take as input a flattened version of the tabular time-series. We select the best linear models after a cross-validated random search of 50 iterations to select their hyper-parameters. Also, regarding the Loan default prediction task, note that only the fine-tuning portion of the Loan dataset is considered for these non-deep learning baselines.
More details on hyper-parameters can be found in Appendix \ref{app:hyperparams}.

\paragraph{Comparability}
To ensure a fair comparison across all models, we implement the same 
pre-training and fine-tuning strategy for all of them. 
While self-supervised pre-training has proved useful on tabular data \cite{rubachev2022revisiting}, we emphasize that our objective is not to design novel pre-training techniques. Thus, we consider a simple field masking pretext task for all models, 
as from previous literature~\cite{padhi2021tabular}.
In detail, we guide pre-training with a BERT-like token masking pretext task \cite{devlin2018bert}, randomly selecting 15\% of the tokens, out of which 80\% are replaced by a \texttt{[MASK]} token, 10\% by a random token and 10\% left unchanged.
Last, as our objective is not to introduce novel fine-tuning mechanisms, we resort to a standard fine-tuning technique popularized in NLP. During fine-tuning, we prepend a \texttt{[CLS]} token to the tabular time-series before encoding it with the model (i.e., before the Final transformer for two-stage models).
Once the full time-series is encoded, this special token is extracted and passed to a final fully connected layer trained on a specific downstream task as in \cite{devlin2018bert}. 
We implement this fine-tuning methodology for all models, with a variation for Tabbie. Indeed, in \cite{iida2021tabbie}, the authors suggest to prepend \texttt{[CLS]} tokens to each row and column processed by Tabbie, based on the downstream task to learn. Thus, we consider the version of Tabbie that yields the best results in our experiments.

\begin{table}[!t]
    \centering
    \fontsize{9}{9}\selectfont
    \caption{Results. Average over 5 seed runs, standard deviation in parenthesis. Models have the same number of parameters.}
    \label{best_results}
    \begin{tabular}{
        @{}l
        @{$\,\,\,$}c 
        @{$\,\,\,$}c 
        @{$\,\,\,$}c
        @{}}
    \toprule
    \textbf{Model} & \textbf{Architecture} &\textbf{\makecell{Pollution\\RMSE $\downarrow$}} & \textbf{\makecell{Loan\\Avg. Precision $\uparrow$}}\\
    \midrule
    Linear & Non-DL & 59.44 {\small(0.28)} & 0.31 {\small(0.03)} \\ 
    XGBoost & Non-DL & 50.74 {\small(0.59)} & 0.36 {\small(0.07)} \\
    \midrule
    FT-Transformer & Single-stage & 26.54 {\small(0.45)} & 0.44 {\small(0.07)} \\
    Tabbie & Single-stage & 22.37 {\small(0.31)} & 0.39 {\small(0.05)} \\
    \midrule
    TabBERT \it{(col-based)} & Two-stage & 26.46 {\small(0.32)} & 0.44 {\small(0.05)} \\
    TabBERT \it{(row-based)} & Two-stage & 21.05 {\small(0.22)} & 0.46 {\small(0.06)} \\
    \sysname{} \it{(ours)} & Two-stage & \textbf{20.13} {\small(0.34)} & \textbf{0.48} {\small(0.06)} \\
    \bottomrule
    \end{tabular}
\end{table}

\subsection{Results}

We evaluate each model over 5 seeds runs and report their average performance and standard deviation in \Cref{best_results}. We emphasize that transformer-based models contain the same total number of parameters. 
On the Pollution dataset, we observe that \sysname{} significantly decreases the RMSE demonstrating the effectiveness of the proposed approach. Regarding the Loan default prediction task instead, we observe that the differences in terms of AP are less significant, likely due to the smaller dataset size. 
Our results for TabBERT (row-based) on the Pollution dataset are similar to the ones reported in \cite{padhi2021tabular,luetto2023one}.
Regarding the Loan default prediction task, given that we reduce the sequence length to any 10 consecutive transactions to generate more fine-tuning samples, we cannot directly compare our results to \cite{luetto2023one} that only used the last 150 transactions instead, at the expense of generating fewer samples to train on. 

We first remark that all transformer models outperform the non-deep learning baselines we evaluate.\footnote{Although prior work demonstrated in many experiments the superiority of gradient-boosted decision trees compared to transformer-based approaches, in our setting, XGBoost might suffer from using quantized numerical features. This is a consequence of our experimental protocol choices which favor comparability with existing literature (especially for transformer-related works) rather than searching for the global best independently.
}
Note that the Pollution prediction task requires to output 20 labels for each input sample, which is implemented with a multi-output regressor wrapped on around these non-deep learning baselines, i.e., fitting one model for each target. Hence, this limits these baselines' ability to capture relationships between targets.
Also, for the Loan default prediction task, the performance gap is partially explained by the pre-training advantage the transformers models are given, as the non-deep learning baselines only use the smaller fine-tuning dataset. 

Among transformer models, single-stage baselines underperform compared to two-stage architectures, highlighting the benefit of hierarchical representations for tabular time-series. In particular, our field-based proposal ranks first on both datasets. In contrast, TabBERT which conditions attention between fields on a unique table axis, yields worse performance indicating that capturing fields relationships across rows \textit{and} columns is important. Additionally, comparing \sysname{} to the flattened FT-Transformer hints at the lack of table structural information for the latter. Last, while Tabbie structures the field embeddings contextualization row-wise and column-wise, its lack of a second-stage fails to relate all of these representations.  

\begin{table}[!t]
\centering
\fontsize{8}{8}\selectfont
    \caption{
    Ablation study.  
    Average over 5 seed runs, standard deviation in parenthesis. 
    \textmd{\emph{Underline} highlights model families best configuration, while \textbf{bold} highlights the global best.}
    }
    \label{ablation}
\begin{tabular}{
@{}l
@{$\,\,\,$}c 
@{$\,\,\,$}c 
@{$\,\,\,$}c
@{$\,\,\,$}c
@{$\,\,\,$}c
@{}
}
\toprule
\textbf{\makecell{Model\\family}} & \textbf{\makecell{Stage with\\more capacity}}  & \textbf{\makecell{Column\\ind. emb.}} & \textbf{\makecell{Row\\pos. emb.}}& \textbf{\makecell{Pollution\\RMSE $\downarrow$}} & \textbf{\makecell{Loan\\AP $\uparrow$}} \\
\midrule
\multirow{4}{*}{FT-Transf.} & \multirow{4}{*}{Single-stage}
    &  &  & 28.28 {\small(0.27)} & 0.43 {\small(0.08)} \\
    &  &  \checkmark &  & 28.04 {\small(0.22)} & 0.42 {\small(0.08)} \\
    &  &  & \checkmark & 27.80 {\small(0.73)} & \emph{0.44} {\small(0.07)} \\
    &  & \checkmark & \checkmark & \emph{26.54} {\small(0.45)} & 0.42 {\small(0.05)} \\
\midrule[1pt]
\multirow{4}{*}{Tabbie} & \multirow{4}{*}{Single-stage} 
    &  &  & \emph{22.37} {\small(0.31)} & 0.38 {\small(0.06)} \\
    &  & \checkmark &  & 22.43 {\small(0.14)} & 0.39 {\small(0.03)} \\
    &  &  & \checkmark & 23.14 {\small(0.23)} & 0.38 {\small(0.03)} \\
    &  & \checkmark & \checkmark & 23.02 {\small(0.14)} & \emph{0.39} {\small(0.05)} \\
\midrule[1pt]
\multirow{8}{5em}{TabBERT \it(col-based)} 
    & \multirow{4}{*}{Field Transf.} 
    &  &  & 27.10 {\small(0.32)} & \emph{0.44} {\small(0.05)} \\
    &   & \checkmark &  & 27.08 {\small(0.32)} & 0.43 {\small(0.04)} \\
    &   &  & \checkmark & \emph{26.46} {\small(0.32)} & 0.40 {\small(0.08)} \\
    &   & \checkmark & \checkmark & 26.48 {\small(0.28)} & 0.42 {\small(0.03)} \\
\cmidrule(lr){3-6}
    & \multirow{4}{*}{Final Transf.}
    &  &  & 27.85 {\small(0.35)} & 0.37 {\small(0.03)} \\
    &   & \checkmark &  & 27.88 {\small(0.30)} & 0.38 {\small(0.02)} \\
    &   &  & \checkmark & 27.19 {\small(0.27)} & 0.38 {\small(0.04)} \\
    &   & \checkmark & \checkmark & 27.23 {\small(0.22)} & 0.36 {\small(0.04)} \\
\midrule[1pt]
\multirow{8}{5em}{TabBERT \it(row-based)} 
    & \multirow{4}{*}{Field Transf.} 
    &  &  & 21.30 {\small(0.28)} & 0.44 {\small(0.04)} \\
    &   & \checkmark &  & 21.07 {\small(0.15)} & \emph{0.46} {\small(0.06)} \\
    &   &  & \checkmark & 21.34 {\small(0.28)} & 0.44 {\small(0.05)} \\
    &   & \checkmark & \checkmark & \emph{21.05} {\small(0.22)} & 0.45 {\small(0.07)} \\
\cmidrule(lr){3-6}
    & \multirow{4}{*}{Final Transf.}
    &  &  & 22.92 {\small(0.32)} & \emph{0.46} {\small(0.05)} \\
    &   & \checkmark &  & 22.72 {\small(0.26)} & 0.44 {\small(0.07)} \\
    &   &  & \checkmark & 22.92 {\small(0.30)} & 0.45 {\small(0.04)} \\
    &   & \checkmark & \checkmark & 22.70 {\small(0.29)} & 0.45 {\small(0.07)} \\    
\midrule[1pt]
\multirow{8}{4em}{\sysname{} \it(ours)} 
    & \multirow{4}{*}{Field Transf.} 
    &  &  & 20.48 {\small(0.19)} & 0.46 {\small(0.07)} \\
    &   & \checkmark &  & \textbf{\emph{20.13}} {\small(0.34)} & 0.44 {\small(0.09)} \\
    &   &  & \checkmark & 20.42 {\small(0.22)} & 0.43 {\small(0.08)} \\
    &   & \checkmark & \checkmark & 20.32 {\small(0.30)} & \textbf{\emph{0.48}} {\small(0.06)} \\
\cmidrule(lr){3-6}
    &  \multirow{4}{*}{Final Transf.} 
    &  &  & 24.15 {\small(0.17)} & 0.41 {\small(0.04)} \\
    &   & \checkmark &  & 24.00 {\small(0.24)} & 0.42 {\small(0.08)} \\
    &   &  & \checkmark & 23.98 {\small(0.22)} & 0.40 {\small(0.06)} \\
    &   & \checkmark & \checkmark & 24.06 {\small(0.27)} & 0.40 {\small(0.06)} \\
\bottomrule
\end{tabular}
\vspace{-1mm}
\end{table}

\subsection{Ablation study}
In \Cref{ablation}, we analyze the effect of various design decisions on the performance of each transformer model on the Pollution and Loan default datasets. \replaceok{Note that, although the majority of the conclusion holds for the two datasets, given the limited number of samples in the Loan default datasets we mainly drive our Ablation study based on the Pollution dataset}{Note that, while the results hold qualitatively for both datasets, quantitative analysis on the Loan task might be affected by the smaller dataset size. In the remainder, we derive conclusions paying more attention to the Pollution dataset results}. 

First, we investigate 
the partition of model capacity between the first stage, i.e., Field transformer, and the second stage, i.e., Final transformer, for hierarchical models. Namely, we modify the number of encoder layers implemented in each stage to either favor one or the other, while ensuring the total number of model parameters remains the same (cf. \Cref{tab:params}).
We observe that hierarchical architectures perform better when favoring the first stage. This is particularly evident for \sysname{} with up to $+16\%$ performance improvement for the Pollution data set. This common trend indicates that the field representations learned in the first stage are particularly important for hierarchical transformers performance. 

Second, we analyze the effect of the table structure encoding mechanisms that we introduced in \Cref{methodology}, namely using column index embedding and/or row position embedding. 
Note that, even for architectures that dissect tabular time-series along table axes (i.e., Tabbie and TabBERT),
the \textit{ordering} of rows and columns is not preserved by design, due to the permutation-invariant nature of the basic attention mechanism, thus requiring additional positional input. Therefore, we evaluate all table structure encoding combinations for all models. 
We observe that explicitly indicating table structure information is beneficial to almost all transformer architectures, compared to not including any. However, the type of structural information required (i.e., column index, row position, or both) is dependent on the model family.
As expected, FT-Transformer benefits from structure information on both axes. Tabbie shows a similar trend on the Loan default prediction task, although differences are less significant. 
Regarding two-stage models, positional encodings are particularly important for the table axis along which field representations are aggregated. Hence, column-based TabBERT benefits from row position information, conversely, row-based TabBERT exploits column index information.
\sysname{} exploits table structure information on both axes, and particularly column index embeddings.

\section{Conclusion and discussion}
\label{discussion}

In this paper we compared transformer architectures for tabular time-series modeling, 
investigating how attention can be used to simultaneously relate tabular fields across rows \textit{and} columns.
We evaluated our approach on tabular regression and classification tasks, showing improvements over existing baselines. 
However, we highlight that this work can be further expanded.
In particular, we envision two possible research directions considering computational costs and more data variety.

\begin{table}[!b]
\fontsize{8}{8}\selectfont
    \centering
    \caption{Attention complexity. $n_{rows}$ and $n_{cols}$ denote the number of rows and columns per tabular time-series.
    }
    \label{complexity}
    \begin{tabular}{ 
    l c c
    }
     \toprule
     \textbf{Model} & \textbf{Stage} & \textbf{Complexity}\\
    \midrule
    FT-Transformer & Single-stage & $\mathcal{O}((n_{rows} \times n_{cols})^2 )$ \\
    \midrule
    Tabbie & Single-stage & $\mathcal{O}({n_{rows}}^2 + {n_{cols}}^2)$ \\
    \midrule
     \multirow{2}{*}{TabBERT \it(col-based)} 
        & 1\textsuperscript{st} -- Field Transf. & $\mathcal{O}({n_{rows}}^2)$ \\
        & 2\textsuperscript{nd} -- Final Transf. & $\mathcal{O}({n_{cols}}^2)$ \\
     \midrule
     \multirow{2}{*}{TabBERT \it(row-based)} 
        & 1\textsuperscript{st} -- Field Transf. & $\mathcal{O}({n_{cols}}^2)$ \\
        & 2\textsuperscript{nd} -- Final Transf. & $\mathcal{O}({n_{rows}}^2)$ \\
     \midrule
     \multirow{2}{*}{\sysname{} \it(ours)} 
        & 1\textsuperscript{st} -- Field Transf. & $\mathcal{O}({n_{rows}}^2 + {n_{cols}}^2)$ \\
        & 2\textsuperscript{nd} -- Final Transf. & $\mathcal{O}((n_{rows} \times n_{cols})^2)$ \\
     \bottomrule
    \end{tabular} 
\end{table}

\paragraph {Computational requirements}
While we ensure that in our experiments all transformer models have the same total number of parameters, their computational requirements differ. In particular, ``vanilla'' self-attention time complexity is $\mathcal{O}(L^2)$ where $L$ is the input sequence length. However, the transformer models we compare do not consider the same input sequence, as they compute attention along various table axes, i.e., row-wise, column-wise or both. 
In \Cref{complexity}, we compare attention complexity for each architecture, based on the number of rows and columns composing the tabular time-series.
We note that the column-based and row-based hierarchical models are equivalent in terms of attention complexity. In contrast, our field-based proposal is essentially combining Tabbie-like attention in the first stage and FT-Transformer attention between all fields in the second stage. Hence, this finer-grained attention comes at a higher computational cost, which translates into longer training and inference time.
Nevertheless, recent techniques for near linear-time attention \cite{xiong2021, shen2021, wang2020} can be readily used to speed up our approach.

\paragraph{Future work}
To further characterize the strengths and weaknesses of  \sysname{}, we require additional evaluations on a more diverse set of tasks and datasets. 
In particular, the size of such datasets require careful consideration, 
given that over-parameterized deep learning models typically tend to overfit small datasets. In this regard, click-through rate data \cite{DiemertMeynet2017} may be of interest for their sequential nature and large volume.
We did not include such dataset in this work because it is not considered in related literature either. Yet, given its larger size, it would be worth considering in a future work.
Also, our findings have only been evaluated in the realm of tabular time-series, however, the wider domain of multi-variate time-series might also benefit from field-based hierarchical architectures, combining length and channel signals in a first stage. 
Finally, as the objective of this paper is limited to the comparison of attention computation axes, we do not include sophisticated tabular data embeddings techniques such as 
numerical features embeddings from \cite{gorishniy2022embeddings, luetto2023one} or target-aware pretext tasks for pre-training from \cite{rubachev2022revisiting}.
We expect their respective benefits to be portable to our novel field-based architecture, as they impact the prior embedding layer of any transformer architecture.
Regarding numerical features embedding, a tangential direction may leverage methods from traditional time-series pre-processing techniques. In this regard, SAX \cite{lin2007experiencing} bins continuous time-series into sequences of discrete symbols to capture their trends. 
Such symbols could then be tokenized before being fed to a transformer-based architecture. 
This motivates an interesting research direction to compare numerical features embeddings schemes in the context of tabular time-series. 

We contribute to the community our codebase and models at \url{https://github.com/raphaaal/fieldy}.

\section{Acknowledgments}
\label{ack}
We would like to thank the authors of \cite{luetto2023one} for fruitful discussions on the success metrics and pre-processing they used.

\normalem
\balance
\bibliographystyle{ACM-Reference-Format}
\bibliography{ref}

\clearpage

\begin{appendices}

\begin{figure}[t!]
\centering
\includegraphics[width=0.80\linewidth]{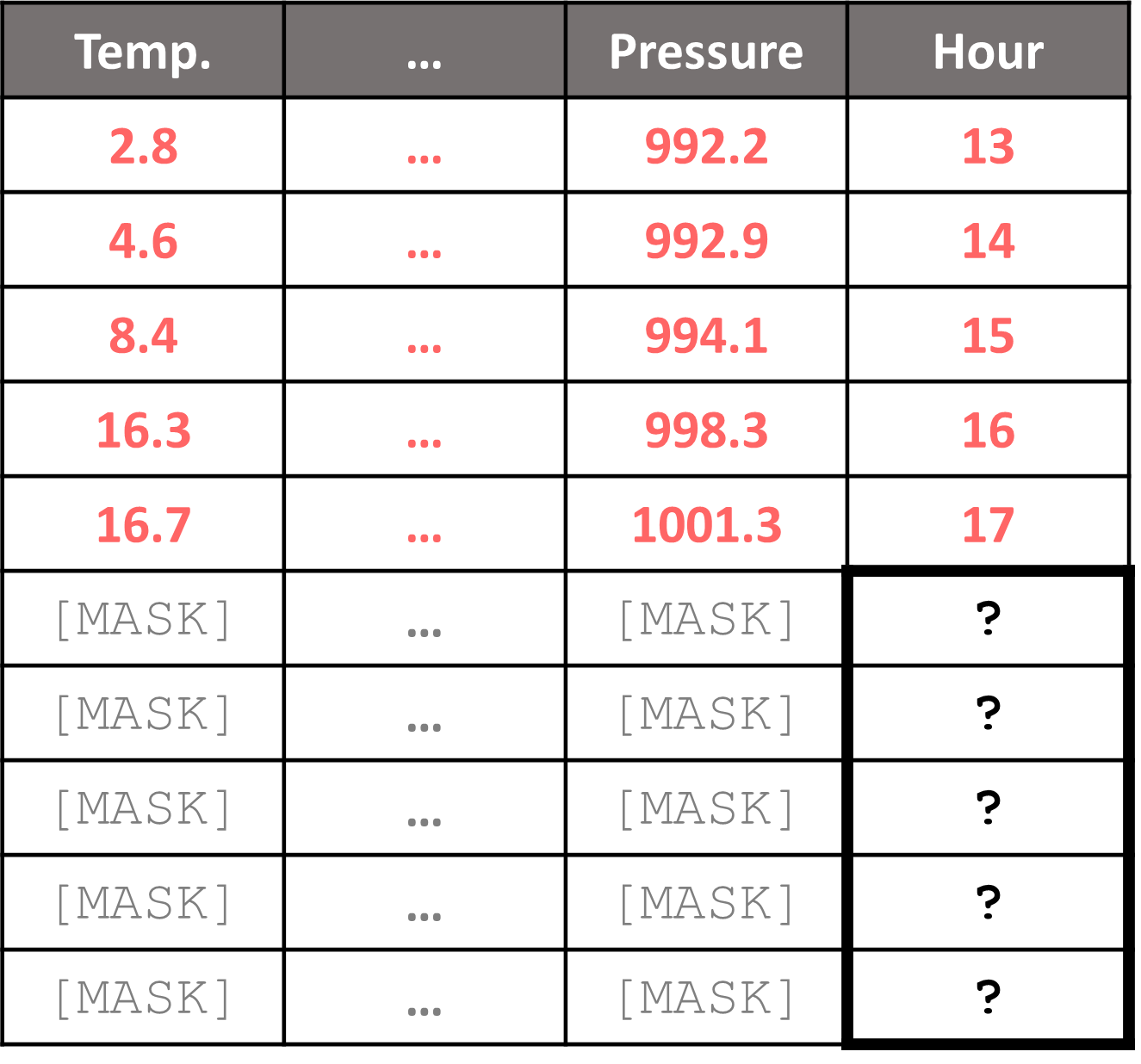}
\caption{Illustrative input sample for the field-wise attention toy task. Predicting missing tokens in the \texttt{Hour} column requires field-wise attention across rows.}
\label{fig:field_attention_investigate}
\end{figure}

\section{Field-wise attention} \label{app:field_attention}
In this section, we investigate the hypothesis that coarse-grained attention resulting from the typical hierarchical learning of tabular time-series representations might miss important cross-field relationships. More precisely, we compare a pre-trained TabBERT (row-based architecture) and a pre-trained \sysname{} (field-based architecture)
on a simple prediction task that specifically requires the attention mechanism to focus on field dependencies \textit{across} rows. For this, we utilize the Pollution dataset introduced in \Cref{experiments}, sampling 100 input sequences of 10 rows each. We then \texttt{[MASK]} all values in the last five rows of each sample, and task the models to predict the masked tokens in the \texttt{Hour} column, as depicted in \Cref{fig:field_attention_investigate}. 
Note that values in the \texttt{Hour} column are always incremented by $+1$, as the Pollution measurements are hourly-based and the sequences are ordered.
Hence, predicting the five missing hour values requires to specifically attend to \texttt{Hour} fields across separate rows.
From this simple experiment, considering the top-1 predicted token,
\sysname{} (field-based) achieves an accuracy of 56\%, significantly outperforming TabBERT (row-based) which only scores 9\%. This indicates that the aggregation performed by hierarchical models such as TabBERT limits their ability to relate fields along their second-stage axis.

\section{Hyper-parameters} \label{app:hyperparams}

Transformers architectures hyper-parameters are reported in \Cref{tab:hyperparams}. Note that we increase dropout and decrease the hidden dimension for the Loan prediction task, as it is composed of fewer samples and models tend to overfit.
We adjust the model size by selecting an appropriate number of layers as reported in \Cref{tab:params}.
We keep all the other hyper-parameters related to model capacity the same across all models for a fair comparison.

Regarding XGBoost, we run random searches to find hyper-parameters, whose possible values are reported in \Cref{tab:xgb}. We use a 2-fold cross-validation on the Pollution dataset and a 10-fold one on the Loan dataset that is significantly smaller. Both datasets were granted a sampling budget of 50 iterations for each seed run.

\begin{table}[ht!]
\centering
\caption{Hyper-parameters for transformer-based models.
}
\label{tab:hyperparams}
\begin{tabular}{lcc}
\toprule
\multicolumn{1}{l}{\textbf{Setup}} & \textbf{Pollution}  & \textbf{Loan} \\ 
\midrule
\multicolumn{1}{l}{Pre-training epochs}& 24 & 60 \\
\multicolumn{1}{l}{Fine-tuning epochs} & 10 & 20 \\
\multicolumn{1}{l}{Optimizer} & AdamW & AdamW \\
\multicolumn{1}{l}{Learning rate} & 5e-05 & 5e-05 \\
\multicolumn{1}{l}{Batch size} & 64 & 100 \\
\multicolumn{1}{l}{Dropout} & 0.1 & 0.3 \\
\multicolumn{1}{l}{Hidden dimension} & 800 & 500 \\
\multicolumn{1}{l}{Number of attention heads} & 10 & 10 \\
\multicolumn{1}{l}{Number of parameters} & $\approx$106M & $\approx$36M \\
\bottomrule  
\end{tabular}
\end{table}

\begin{table}[h]
\centering
\caption{Transformer-based models layers statistics.
}
\label{tab:params}
\begin{tabular}{lc r@{$\,$}l r@{$\,$}l}
\toprule
\textbf{Model} & \bf Stage with & \multicolumn{4}{c}{\bf Num. of layers} \\
\textbf{family} & \bf more capacity & \multicolumn{2}{c}{\bf Pollution}  & \multicolumn{2}{c}{\bf Loan} \\ 
\midrule
FT-Transformer & Single-stage & \multicolumn{2}{c}{14} & \multicolumn{2}{c}{8} \\
\midrule
Tabbie & Single-stage & \multicolumn{2}{c}{4} & \multicolumn{2}{c}{4} \\
\midrule
\multirow{2}{*}{TabBERT \textit{(col-based)}} 
    & Field Transf.& \multicolumn{2}{c}{6 / 10} & 6 & / 6 \\
    & Final Transf. & \multicolumn{2}{c}{1 / 12} & 1 &/ 8 \\
    
\midrule
\multirow{2}{*}{TabBERT \textit{(row-based)}} 
    & Field Transf. & \multicolumn{2}{c}{6 / 10} & 6 &/ 6 \\
    & Final Transf. & \multicolumn{2}{c}{1 / 12} & 1 &/ 8 \\
    
\midrule
\multirow{2}{*}{\sysname{} \textit{(ours)}} 
    & Field Transf. & \multicolumn{2}{c}{8 / 4} & 5 &/ 4 \\
    & Final Transf. & \multicolumn{2}{c}{2 / 10} & 2 &/ 6 \\
\bottomrule  
\end{tabular}
\\
\raggedright
\small Parameter count per layer depends on the architecture. For hierarchical architectures, values reflect 1\textsuperscript{st}-stage layers / 2\textsuperscript{nd}-stage layers. 
\end{table}

\begin{table}[b]
\centering
\caption{Hyper-parameters for XGBoost random search.}
\label{tab:xgb}
\begin{tabular}{
   @{}l
    @{$\,\,\,$}c 
    @{$\,\,\,$}c 
    @{}
}
\toprule
\textbf{Parameter} & \textbf{\makecell{Pollution}}  & \textbf{\makecell{Loan}} \\ 
\midrule
Objective & MSE & Logistic \\
Max \# trees & 5,000 & 5,000 \\
Early-stopping & 50 & 50 \\
Max depth & [1, 2, ..., 20, None] & [1, 2, ..., 20, None] \\
Learning rate & LogUniform(1e-05, 0.7) & LogUniform(1e-05, 0.7) \\
Min child weight & LogUniform(1e-08, 100) & LogUniform(1e-08, 100) \\
Subsample & Uniform(0.5, 1) & Uniform(0.5, 1) \\
Col. sample & Uniform(0.5, 1) & Uniform(0.5, 1) \\
Gamma & LogUniform(1e-08, 100) & LogUniform(1e-08, 100) \\
Reg. alpha & LogUniform(1e-08, 100) & LogUniform(1e-08, 100) \\
Reg. lambda & LogUniform(1e-08, 100) & LogUniform(1e-08, 100) \\
\bottomrule  
\end{tabular}
\end{table}

\clearpage

\section{Datasets} \label{app:datasets}

In \Cref{tab:data}, we describe the two datasets used for evaluation. For the Pollution task, the same dataset is used for pre-training (unlabeled) and fine-tuning. For the Loan default prediction task, the pre-training dataset is composed of 4,500 clients with 232 transactions on average. At each pre-training step, random sequences of 10 consecutive client transactions are sampled for masking. The fine-tuning dataset for this task corresponds to transactions from 682 clients. As in prior work \cite{luetto2023one}, splitting between training, validation and test sets is performed on a client-basis to avoid target leakage.

\balance

\begin{table}[h]
\centering
\caption{Datasets statistics.} 
\label{tab:data}
\begin{tabular}{lcc}
\toprule
\textbf{Item} & \textbf{Pollution} & \textbf{Loan} \\ 
\midrule
Time-series \# rows & 10 & 10 \\
Time-series \# columns & 16 & 10 \\
\# categorical columns & 7 & 7 \\
\# numerical columns & 9 & 3 \\
\midrule
Pre-training samples & 67K & $\approx$999K \\
Pre-training split (train-val-test) & 60-20-20 & 80-20-0 \\
\midrule
Fine-tuning samples & 67K & 5K \\
Fine-tuning split (train-val-test) & 60-20-20 & 60-20-20 \\
\bottomrule    
\end{tabular}
\end{table}

\clearpage

\end{appendices}


\end{document}